\relax
\documentclass[letterpaper]{article} 
\usepackage{aaai22}  
\usepackage{times}  
\usepackage{helvet}  
\usepackage{courier}  
\usepackage[hyphens]{url}  
\usepackage{graphicx} 
\usepackage{natbib}  
\usepackage{caption} 
\DeclareCaptionStyle{ruled}{labelfont=normalfont,labelsep=colon,strut=off} 
\usepackage{latexsym}
\usepackage{amssymb}
\usepackage{amsmath}
\usepackage{amsthm}
\usepackage{booktabs}
\usepackage{enumitem}
\usepackage[T1]{fontenc}
\usepackage[utf8]{inputenc}
\usepackage{microtype}
\usepackage{inconsolata}
\usepackage{xcolor}
\usepackage{mdframed}
\usepackage{array}
\usepackage{pifont}
\usepackage[linesnumbered,ruled,vlined]{algorithm2e}
\usepackage{hyperref} 

\newcommand{\BibTeX}{B\kern-.05em{\sc i\kern-.025em b}\kern-.08em\TeX}

\title{Instruction Following with Goal-Conditioned Reinforcement Learning in Virtual Environments}

\author{
Zoya Volovikova,\textsuperscript{\rm 1,\rm 2}
Alexey Skrynnik,\textsuperscript{\rm 1,\rm 3}
Petr Kuderov,\textsuperscript{\rm 1,\rm 2}
Aleksandr I. Panov\textsuperscript{\rm 1,\rm 2,\rm 3}\\
}
\affiliations {
\textsuperscript{\rm 1} AIRI, Moscow, Russia\\
\textsuperscript{\rm 2} MIPT, Moscow, Russia\\
\textsuperscript{\rm 3} FRC CSC RAS, Moscow, Russia\\
volovikova@airi.net
}

\begin{document}

\maketitle

\begin{abstract}
    In this study, we address the issue of enabling an artificial intelligence agent to execute complex language instructions within virtual environments. In our framework, we assume that these instructions involve intricate linguistic structures and multiple interdependent tasks that must be navigated successfully to achieve the desired outcomes. To effectively manage these complexities, we propose a hierarchical framework that combines the deep language comprehension of large language models with the adaptive action-execution capabilities of reinforcement learning agents. The language module (based on LLM) translates the language instruction into a high-level action plan, which is then executed by a pre-trained reinforcement learning agent. We have demonstrated the effectiveness of our approach in two different environments: in IGLU, where agents are instructed to build structures, and in Crafter, where agents perform tasks and interact with objects in the surrounding environment according to language commands.
\end{abstract}

    \section{Introduction}

    The ability to solve complex tasks, formulated in natural language, that require a long sequence of actions in the environment is a fundamental property of human intelligence. The recent significant success of large language models (LLMs) in instruction following and explanation generation demonstrates their powerful capabilities in solving commonsense, general knowledge, and code generation problems within the verbal domain. However, success rate of multi-step task completion for autonomous agents driven by general purpose LLMs is still low~\cite{liu2024agentbench}. Moreover, LLMs are often trained solely on textual data, which limits their ability to understand and perform actions in real-world-like environments. Consequently, even ChatGPT~\cite{ouyang2022training} exhibits poor spatial reasoning~\cite{bang2023multitask}. On the other hand, reinforcement learning (RL) has proven effective in learning sequences of fine-grained actions for specific tasks within an environment. Thus, investigating the combination of LLMs for natural language understanding and high-level planning, along with RL for learning environmental manipulation, represents a promising research direction.

    \begin{figure}[t]
        \centering
        \includegraphics[width=1.0\linewidth]{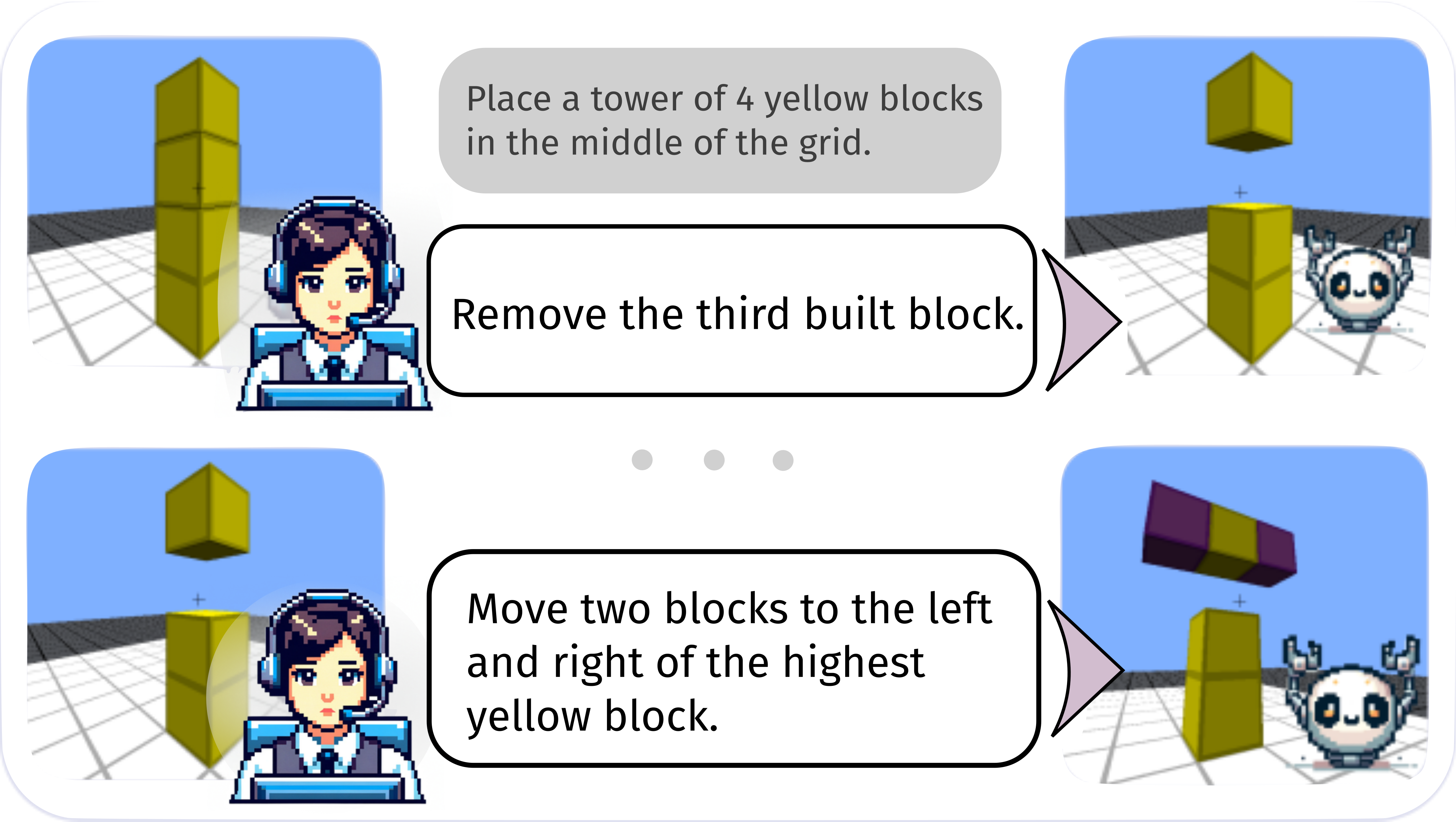}
        \vspace*{-10px}
        \caption{The task of collaborative interaction between the agent, the environment, and the user involves the following: the user provides instructions to the agent, and the agent executes actions within the environment to accomplish the task based on these instructions.}
        \label{fig:igluset}
     \vspace*{-6px}
    \end{figure}

    LLMs can be regarded as universal knowledge bases that allow human users to interact in natural language and solve complex tasks~\cite{huang2022language, zeng2022socratic}. Recent studies have shown that pre-trained LLMs can construct high-level action plans in both simulated and real environments~\cite{bara2021mindcraft,min2022film,murray2022following}. However, these LLM-based approaches necessitate manual prompt engineering, handcrafted translation of language commands into embodied actions, and a strategy for goal-aware action selection from the distribution of potential options generated by language. In this context, several studies~\cite{ahn2022can,ouyang2022training} have demonstrated that the rough action plans extracted from language models can be refined using RL.

    In our research, we present the hierarchical framework \textbf {IGOR (Instruction Following with Goal-Conditioned RL)}, which combines the capabilities of large language models for understanding complex natural language constructions with RL-based policies to develop effective behavior in the embodied environment. The framework relies on two main components: a \textbf {Language Module}, which translates instructions in natural language into a high-level action plan with generated subgoals, and a \textbf {Policy Module}, tasked with executing this plan and achieving subgoals.

    Furthermore, we introduce independent learning strategies for each module. We have developed efficient learning strategies for the LLM for limited datasets. Some of these strategies are based on data augmentation using the ChatGPT model, while others rely on subdividing subtasks by altering data formats and decomposing these subtasks into primitives. We set a learning task for the RL agents based on goals and curriculum. This approach promotes highly efficient task execution in dynamic environments beyond the training sample.

    The effectiveness of our approach was rigorously tested in two embodied environments. The first is IGLU~\cite{Zholus2022b}, where the 'Builder' agent constructs structures based on natural language instructions from the 'Architect' agent (see an example in Fig.~\ref{fig:igluset}). The second is the Modified Crafter environment, where the instruction-following agent needs to adapt to dynamically changing environments. Our results demonstrate that the application of our method not only outperforms the algorithms presented in the NeurIPS IGLU competition\footnote{\href{https://www.iglu-contest.net}{https://www.iglu-contest.net}}~\cite{kiseleva2022iglu, kiseleva2022interactive, kiseleva2023interactive}, but also surpasses the baselines based on Dreamer-v3 in Crafter environment~\cite{hafner2022benchmarking}.
    
    The \textbf{main contributions}\footnote{Our code is available at \href{https://github.com/AIRI-Institute/IGOR}{https://github.com/AIRI-Institute/IGOR}} of our paper are: 
    \begin{enumerate}
        \item We proposed a novel task decomposition approach that facilitates the incorporation of augmentations, curriculum learning, and potentially other techniques to multi-modal setups involving LLM and RL learning for virtual environments.
        \item In addition to the known IGLU environment, we presented a modified Crafter environment by introducing a textual modality and prepared a corresponding dataset to support this enhancement.
        \item We conducted extensive experiments to compare our approach with other methods in both the Crafter and IGLU environments, demonstrating significant improvements.
    \end{enumerate}

    \section{Related work}

    \textbf{Planing with LLM.} Recent studies are actively exploring the generation of action plans using language models. Some works focus on prompt engineering for effective plan generation~\cite{singh2022progprompt}, while others address the challenge of translating the language model's output into executable actions~\cite{huang2022language}. In~\cite{ahn2022can}, models are trained with RL for selecting feasible actions and executing elementary actions. Many of these works~\cite{shridhar2020alfred, lynch2023interactive} aim to control robots interactively in real time. Maintaining a dialogue with humans is an essential area of research for robotics~\cite{padmakumar2022teach, gao2022dialfred}.
    
    \textbf{Language Grounding Problem.} The language grounding problem in intelligent agents involves linking objects across modalities, such as matching textual instructions with objects in virtual environments. Methods to address this include using CLIP for visual-textual links~\cite{Clip}, cross-attention mechanisms, and compressing data into hidden subspaces, exemplified by Dynalang~\cite{Dynalang}. Some strategies integrate language processing with reinforcement learning, using text embeddings as observations~\cite{DreamerV3, Dynalang}. Others connect textual descriptions to environmental entities using transformer models like Emma and EmBERT~\cite{Emma, EmBert}. Additionally, some approaches use multiple modules trained independently, with pre-trained language models aiding in planning and adapting actions, addressing the lack of real-world experience~\cite{huang2022language, brohan2023can, li2022pre}. Innovatively, models like JARVIS-1 combine pre-trained memory blocks with tools like CLIP, enhancing multimodal memory and scheduling~\cite{jarvis1, fan2022minedojo}.

    \textbf{Embodied environments.} In the field of embodied reinforcement learning, several platforms have been developed to train agents based on text instructions. Among these, AI2Thor~\cite{Ai2-thor} and Habitat~\cite{szot2021habitat} , offer tasks that are simple and adhere to strict rules, which simplifies the process of linking actions to text using straightforward syntactic structures (Messenger~\cite{Emma}, HomeGrid~\cite{Dynalang}, TWOSOME~\cite{TWOSOME}).

    Furthermore, advancements have been made to enhance the Crafter~\cite{hafner2022benchmarking}  environment, resulting in the creation of the Text-Crafter~\cite{du2023guiding} version. Similarly, the MineDojo~\cite{fan2022minedojo} platform, which is based on Minecraft, has been introduced. These platforms are designed for more intricate linguistic and planning tasks. Additionally, the IGLU~\cite{kiseleva2023interactive} environment stands out for its complexity. In IGLU, agents must follow detailed instructions to construct structures within a virtual world. These environments are characterized by a vast state space and involve complex tasks that are formulated by humans.

    \section{IGOR: Follow Instruction with Goal-based RL}

    \begin{figure}[!ht]
        \vspace*{-5px}
        \centering
        \includegraphics[width=0.85\linewidth]{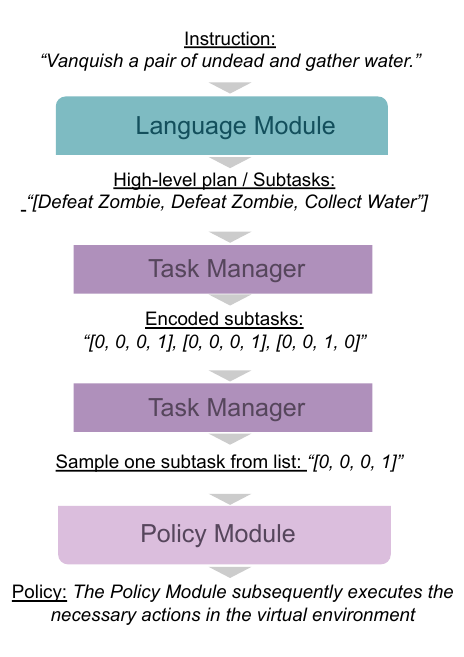}
        \vspace*{-10px}
        \caption{The IGOR framework has three modules: a Language module that solves language understanding problems and provides a high-level plan of subtasks, a Task Manager that encodes the subtasks for the Policy module, and a Policy module that executes actions in the environment based on visual observations and subtasks.}
        \label{fig:flow_example}
        \vspace*{-10px}
    \end{figure}

    \begin{figure*}[!ht]
        \centering
        \includegraphics[width=1.0\linewidth]{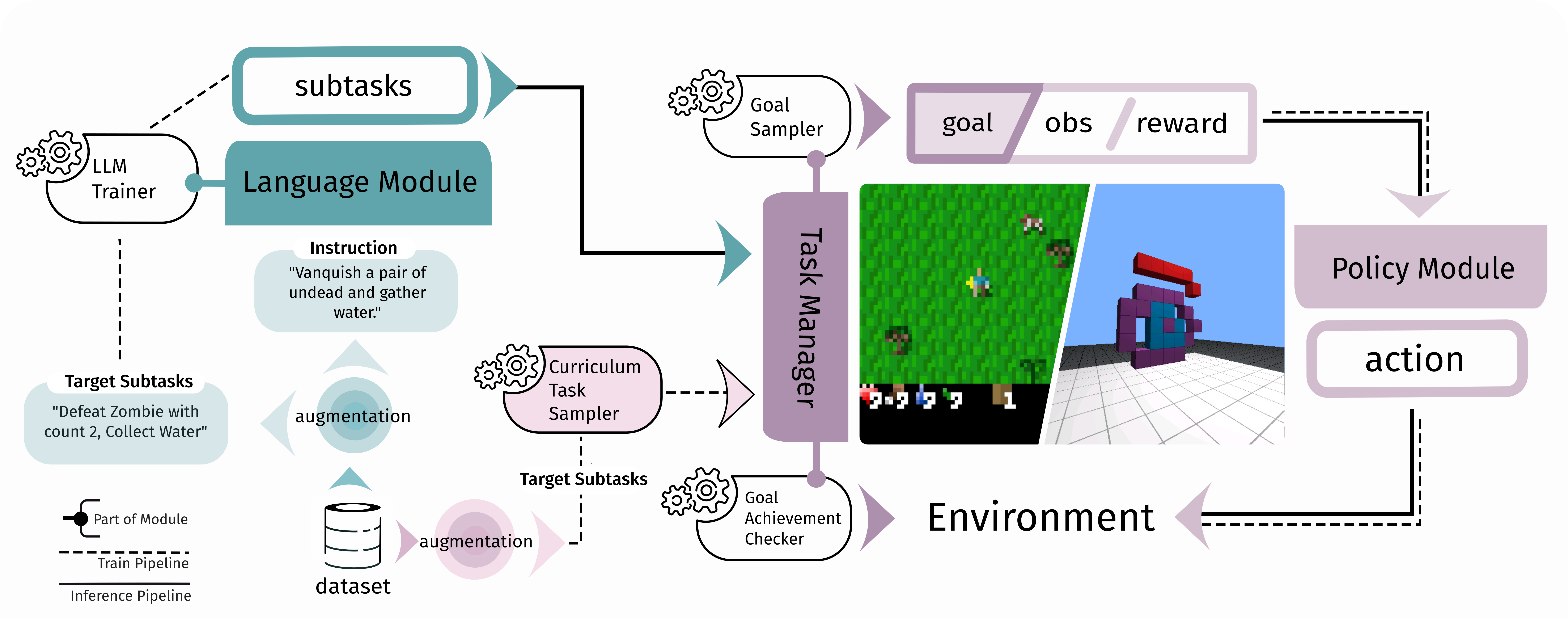}
        \vspace*{-18px}
        \caption{The diagram displays the IGOR system, where the "Language Module" transforms text instructions into subtasks. The "Task Manager" coordinates the subtasks and monitors their execution. The "Policy Module" operates in a virtual environment based on the subtasks. Dotted lines indicate the training process of the modules, while solid lines show how the modules interact during inference.}
        \label{fig:IGORscheme}
        \vspace*{-10px}
    \end{figure*}

    The IGOR framework is designed to solve the challenges of natural language understanding and instruction execution within virtual environments, enabling the processing of instructions that contain complex linguistic structures and specific terminologies. The virtual environments in which the intelligent agent operates are characterized by extensive observation areas and require the execution of multiple interconnected tasks. The framework is composed of three key modules.

    The \textbf {Language Module}, implemented using Large Language Models (LLM), analyzes incoming instructions and converts them into a high-level plan consisting of a set of specific subtasks that need to be executed.

    The \textbf {Policy Module}, implemented using reinforcement learning methods based on the Proximal Policy Optimization (PPO) algorithm, is responsible for the strategic interaction in the environment, including the execution of the interconnected tasks.

    The \textbf{Task Manager} acts as a wrapper over the virtual environment, transforming the list of subtasks provided by the Language Module into a format understandable to the Policy Module. This module also ensures that the tasks specified in the instructions are completed and concludes the episode after their execution.

    Thus, the \textbf{inference pipeline} operates by initially receiving instructions which are processed by the Language Module, where they are translated into a series of subtasks --- a high-level execution plan. Subsequently, the Policy Module executes the necessary actions in the virtual environmrnt to achieve the objectives outlined in the instructions. An example of using our pipeline can be found in the Figure ( \ref{fig:flow_example})

    \textbf {Training} for each learnable module is conducted separately, which provides flexibility in the integration of training methods and techniques. The training of the Language Module involves various augmentations and modifications to the dataset to prevent overfitting. The Policy Module is trained using a goal-based approach, which allows for training on a broader set of potential objectives than those available in the initial data. The inclusion of a curriculum in the training process also significantly enhances the quality of the final agent.

    \subsection {Language Module Training Techniques}\label{llm_str}

    The core of our language module utilizes a large pre-trained language model, which has been further finetuned on a dataset specific to the environment. This dataset contains information on how instructions translate into specific subtasks, enabling the model to understand and decompose complex commands effectively. The training leverages a specific format that maps instructions to their corresponding subtasks (e.g., Instruction -> Subtask 1, Subtask 2, Subtask 3).

    Due to the difficulty of obtaining comprehensive datasets for training models to translate language instructions into commands, we face additional challenges. Typically, datasets for training large language models (LLMs) on such tasks are manually curated, which is a labor-intensive process. This often limits both the size and the quality of the datasets available. In response, we employ techniques to prevent overfitting, especially when working with these limited datasets. Our experiments demonstrate that these methods effectively enhance training quality by ensuring the model can generalize well from smaller, varied linguistic datasets, leading to a more robust understanding of instructions.

    \textbf{Augmentation with LLM}. In this technique, we begin by understanding the structure and specific terminology of the dataset. Our approach involves iteratively modifying the list of subtasks necessary to execute a given instruction for each dataset element. Subsequently, ChatGPT or another LLM is tasked with rewriting the instruction to incorporate these modified subtasks. 

    The prompt request is structured as follows:

    \textit{1) Description of the Environment:} Provide the LLM with a detailed understanding of the setting by explaining the overarching themes and key specific concepts. Use succinct descriptions for familiar ideas and detailed explanations for unique aspects.

    \textit{2) Few-shot Example:} Introduce the original instruction alongside its required subtasks. Optionally, you can also provide an example of how the instruction might change if the subtasks are altered.

    \textit{3) Task Modification Request:} Specify new subtasks for the target instruction and request the LLM to revise the instruction accordingly.

    It is important to emphasize that the LLM is not creating the instruction from scratch. We aim to start with the existing instruction and suggest modifications, ensuring that the style of the original instruction is preserved as much as possible.

    \textbf{Subtasks decomposition.} The second technique entails decomposing original subtasks into "primitives", which involves modifying the structure and format of subtasks in the dataset. Essentially, it suggests consolidating frequently co-occurring subtasks into a single common subtask, thereby reducing the data volume required for processing by the language model. These aggregated subtasks are termed "primitives".

    We explore two methods for creating primitives. One method utilizes unique tokens, such as emojis, to encode each primitive. Emojis are chosen for their diverse range, making them a convenient means of representing a broad array of subtasks. The second method involves crafting primitives in a manner that aligns logically with the content of the instructions. This approach aims to enhance the coherence between the instructional context and the subtask structure and is the approach employed in our experiments.

\subsection {Task Manager}

The Task Manager serves as an intermediary to connect the Language Module and the Policy Module during execution and to allocate subtasks from the dataset to the Policy Module throughout its training phase. It retrieves a list of subtasks and systematically supplies these to the RL agent as components of its observational input. Based on observations collected during the agent's interaction with the environment, the Task Manager determines whether a subtask has been successfully completed and decides whether to proceed to the next task or to conclude the episode. Upon successful completion of a subtask, the Task Manager assigns a positive reward ($r= +1$) to reinforce the agent's behavior.

\subsection {Training the Policy Module}

 Instead of training a RL agent to tackle all subtasks required by a complex instruction simultaneously, we propose a \textbf{goal-based} learning approach. In this approach, each episode presents the agent with an observation and only one of the subtasks from the instruction. By doing this, we simplify the agent's task, reducing complexity of the task and facilitating more focused learning on specific aspects of the overall problem.

We examine a visual-based reinforcement learning environments characterized as a Partially Observable Markov Decision Process (POMDP). In this framework, the observation function is augmented by subtask encodings provided by the Task Manager. The primary objective of the agent is to develop a policy \(\pi\) that selects actions to maximize the expected cumulative reward. To achieve this, we employ policy gradient methods, specifically the Proximal Policy Optimization (PPO) algorithm. PPO has been demonstrated to offer substantial robustness across various tasks, attributed to its effective balancing of exploration and exploitation by optimizing a clipped surrogate objective function \(\mathbb{E}\left[\min(\rho_t(\theta) \hat{A}_t, \text{clip}(\rho_t(\theta), 1-\epsilon, 1+\epsilon) \hat{A}_t)\right]\), where \(\rho_t(\theta) = \frac{\pi_\theta(a_t|s_t)}{\pi_{\text{old}}(a_t|s_t)}\) and \(\hat{A}_t\) denotes the advantage estimate at time \(t\)~\cite{schulman2017proximal}.

During training we utilize curriculum task sampler that dynamically adjusts the probability of selecting subtasks based on their performance, inspired by curriculum learning approaches~\cite{matiisen2019teacher, nesterova2022reinforcement}. Specifically, the selection probability for each task $i$ in $\mathcal{T}$ is modified according to:
\[
q_i = 
\begin{cases} 
\frac{1}{d} & \text{if } r_i \geq \tau \\
1 + (\delta_i \cdot d) & \text{if } r_i < \tau 
\end{cases}
\]
where $r_i$ is the task's average reward, $\delta_i$ measures the variability in reward, $d$ is a scaling coefficient, and $\tau$ is a success threshold. Probabilities are normalized using a softmax function to form a distribution from which tasks are sampled. A detailed description of the algorithm, its pseudocode and ablation study can be found in the Appendix.

\section{Experimental Setup}

To investigate and test the capabilities of intelligent agents, we have chosen environments with high combinatorial complexity. These environments allow us to assess how agents cope with tasks requiring the execution of many interrelated subtasks to achieve a target state.

The first environment is IGLU, where the agent's task is to build three-dimensional constructions based on textual descriptions. It is important to note that the complexity of the environment largely lies in the fact that, depending on the instructions, there can be a vast number of potential target states. To be successful in such an environment, an agent must possess advanced text interpretation skills, as well as the ability to think spatially and model ordering to adequately recreate the required structures.

The second environment is Crafter, where the agent needs to follow textual instructions to perform a variety of tasks, such as gathering resources and crafting items. This environment tests the agent's ability to understand natural language and effectively plan sequences of actions in response to changing conditions.

Below is the general pipeline for applying our approach to these virtual environments:

\begin{enumerate}
    \item \textbf{Fine-tune the LLM}: Fine-tune the large language model (LLM) with an environment-specific dataset to translate instructions into subtasks for the reinforcement learning (RL) agent. Add environment-specific techniques if needed: augmentation, data modification, and dataset expansion.
    \item \textbf{Train the RL agent}: Train the RL agent in a goal-based mode on environment-relevant subtasks. If needed, add environment-specific techniques: curriculum learning, hierarchical RL, reward shaping, and curiosity-driven exploration.
    \item \textbf{Measure performance}: Measure performance on a test dataset for each environment. The LLM predicts subtasks from instructions, which the trained RL agent then executes.
\end{enumerate}

\subsection{IGLU Environment}

\paragraph{Environment.} IGLU is an environment\footnote{\href{https://github.com/iglu-contest/gridworld}{https://github.com/iglu-contest/gridworld}} where an embodied agent can build spatial structures (figures) of blocks of different colors. The agent's goal is to complete a task expressed as an instruction written in natural language.

\begin{figure}[ht!]
\centering
    \includegraphics[width=0.32\linewidth]{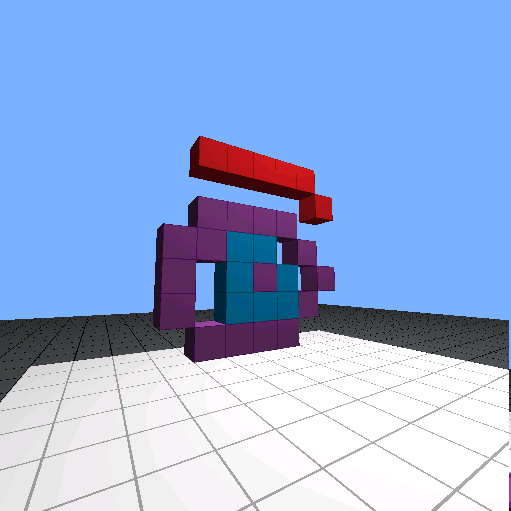}
    \includegraphics[width=0.32\linewidth]{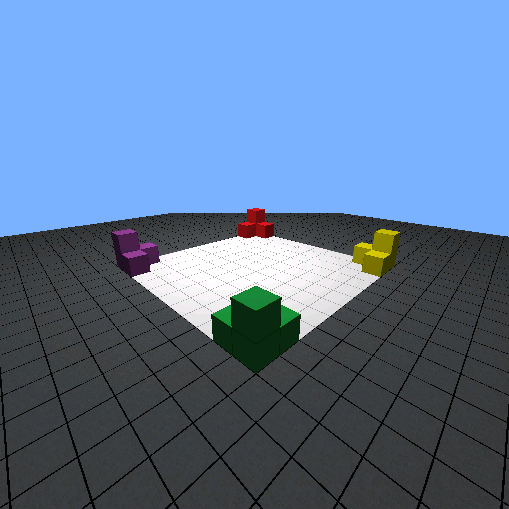}
    \includegraphics[width=0.32\linewidth]{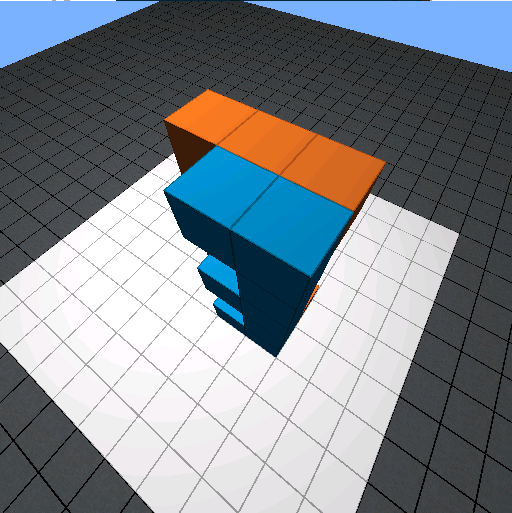}
    \caption{IGLU is a 3D environment where agents are tasked with constructing structures in a designated area, guided by descriptions provided in natural language and the agent’s first person perspective.}
    \label{fig:iglu-examples}
\end{figure}

The observation space consists of point-of-view (POV) image $(64, 64, 3)$, inventory item counts $(6)$, and the pitch and yaw angles $(5)$. The agent can navigate over the building zone, place, and break blocks, and switch between block types. Additionally, the environment provides a textual observation in the form of a dialogue from the dataset, which defines a building task. The examples of such target tasks is presented in Fig.~\ref{fig:iglu-examples}.

Target utterances define the rest of the blocks needed to be added. The environment provides two modes of action for choice: walking and flying. In our experiments, we create agents for use in both flying and walking modes. The action space combines discrete and continuous subspaces: a discrete space of 13 actions (\textit{noop, four camera rotation actions, break block, place block, choose block type 1-6}) and a set of 6 continuous actions for movement in all three directions.

\paragraph{Metrics.} We employed the \textit{F1} metric, as proposed in the IGLU competition, to evaluate the quality of the approaches. 

\paragraph{Baselines.}
We have selected the three best solutions from the IGLU 2022 competition for comparison. The second and third-place solutions, which utilized T5~\cite{raffel2020exploring} and Pegasus~\cite{zhang2020pegasus} models respectively, are based on the organizers' proposed solution but differ in the NLP models employed. We also include the solution of the first-place team, called BrainAgent\footnote{\href{https://github.com/kakaobrain/brain-agent}{https://github.com/kakaobrain/brain-agent}}, whose approach differs significantly from the others. They used end-to-end learning, with the RL agent taking the embedding of a frozen NLP model as input, along with environment information and manually added features.

\paragraph{Dataset.} The training and testing datasets contain 109 and 41 English instructions, respectively, with corresponding grids that need to be constructed. Each instruction consists of individual building steps with a corresponding voxel for each step.

For training the language model, we transform voxel shapes into text --- a list of block coordinates that need to be placed. Grids are converted into a sequence of block coordinates and colors in the form of a string of tuples. For this dataset, a single block (its coordinates and a color) represents a single subtask in our terminology.

The order of the blocks in the sequence corresponds to the appearance of the blocks in the grid and have the following ordering rule: first by $x$, then by $z$, and then by $y$, where $x$ and $z$ are horizontal axes and $y$ is vertical axis. In many instructions, the horizontal position of the first block is not specified, such that it can appear anywhere along $x$ and $z$ randomly. This is detrimental to the model during training as it will lead to high loss values for tokens that cannot be derived from the input data. Therefore, we change the sequence so that the first block of the first step of the instruction is placed at the center of the space, and the other blocks are shifted accordingly.

\textit{Figures Classification}

Based on the capabilities an AI agent needs to construct figures in a dataset, we have developed our own rule-based classification of figures. This helps to more precisely determine which types of figures pose challenges and which descriptions of figure types might cause difficulties in construction by a language model. For visual examples, see Appendix. The classes, defined by their spatial characteristics, are presented below:

\begin{itemize}
\item \texttt{flat} figures are characterized primarily by their length and width, with minimal emphasis on height.
\item \texttt{tall} figures prioritize height over both length and width.
\item \texttt{air} figures are distinguished by their lack of contact with the ground, signifying that they are airborne.
\item \texttt{floor} figures are defined by their complete contact with the ground, resting entirely on the surface.
\end{itemize}

\paragraph{Training details.} The \textit{Language Module} for the IGLU environment was trained to predict, based on dialogue instructions, a list of coordinates of blocks that need to be placed or removed, where each coordinate corresponds to a subtask.We created a second dataset with a modified format of subtasks, grouping blocks that occur together in instructions into parallelepipeds. An example of the subtasks format for the IGLU dataset can be found in \ref{tab:iglu_format}. Additionally, we augmented the dataset using ChatGPT, where we tasked the model to rotate the object by 180 degrees. The modification we applied for ChatGPT augmentation involved rotating the object by 180 degrees. We requested the LLM to adjust the building instructions to match this modification, ensuring the textual description aligned with the altered figure orientation.

As the base for the Language Module, the Flan-T5 base model was used. The LLM models are trained in a seq2seq manner, converting the instruction into a sequence of blocks as described above. Training hyperparameters and hardware parameters are listed in the appendix \ref{ap-params}.  

\begin{table}[!ht]

\caption{Example of \textbf{coords} and \textbf{prims} subtask formats in the IGLU environment, corresponding to the instruction: \textit{"<Architect> place 5 red blocks in a row, one row north of center."}}
  \centering
  \fontsize{7.5}{8}\selectfont 
  \begin{tabular}{@{}lp{3.3cm}p{3.3cm}@{}}
    \toprule
    \textbf{Format} & \textbf{Description} & \textbf{Example} \\ \midrule
    
    \textbf{coords} & 
    (x, y, z, colorID):
    \begin{itemize}[left=0pt, itemsep=0pt] 
      \item \textbf{x, y, z}: coordinates
      \item \textbf{colorID}: block color id
    \end{itemize} & 
    \textit{(0, 5, 5, 3), (0, 6, 5, 3), (0, 7, 5, 3), (0, 8, 5, 3), (0, 9, 5, 3)} \\ 
    \midrule
    
    \textbf{prims} & 
    (start), (size), rotation, color:
    \begin{itemize}[left=0pt, itemsep=0pt] 
      \item \textbf{(start)}: initial block (x, y, z)
      \item \textbf{(size)}: dimensions (x, y, z)
      \item \textbf{rotation}: alignment
      \item \textbf{color}: block color name
    \end{itemize}& 
    \textit{(0, 5, 5), (1, 1, 5), eastsky, red} \\ \bottomrule
  \end{tabular}
  
  \label{tab:iglu_format}

\end{table}

For training \textit{Policy Module} in the IGLU  environment, our objective is to train the agent to either place or remove a single block on a virtual field. At the beginning of an episode, a list of subtasks is randomly generated or taken from a training dataset. A specific block from this list is then selected as the target. Depending on this selection, we generate the initial state of the environment by positioning the blocks leading up to the target block.

The reward mechanism used during training is the same as proposed by the competition's authors.

To enhance the quality of training for the RL agent, we incorporate a curriculum that controls the complexity of the list of subtasks. This structured approach helps in progressively challenging the agent, thus improving its learning efficiency and capability to handle complex tasks.

Solving visual goal-based tasks can be accomplished by a wide range of modern RL approaches, both model-free~\cite{berner2019dota, espeholt2018impala, petrenko2020sample} and model-based~\cite{hafner2022benchmarking, DreamerV3}. In this work, we used an open-source implementation of asynchronous PPO\footnote{https://github.com/alex-petrenko/sample-factory}~\citep{petrenko2020sample}, which demonstrated better performance in terms of GPU hours to final score based on our preliminary experiments. 
\subsection{Crafter Environment}

\begin{figure}[ht!]
\centering
\includegraphics[width=0.32\linewidth]{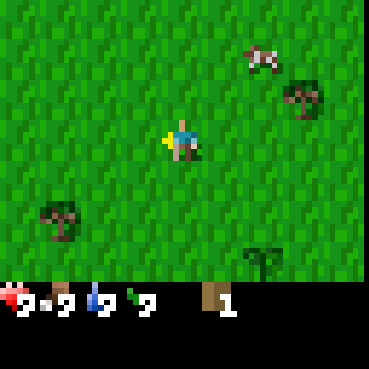}
\includegraphics[width=0.32\linewidth]{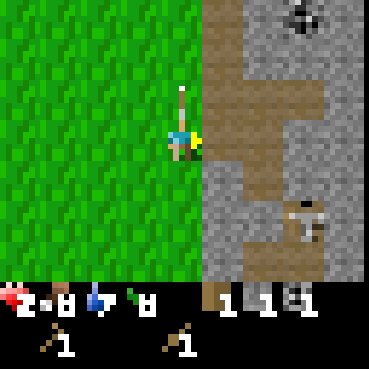}
 \includegraphics[width=0.32\linewidth]{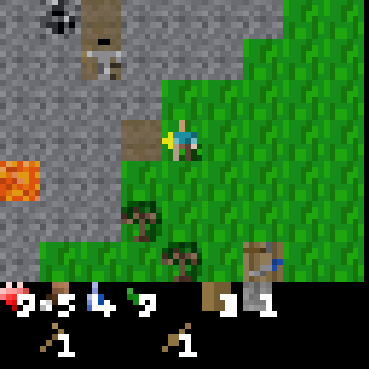}
\caption{Crafter is a 2D environment reminiscent of Minecraft, where players must gather food and water, acquire resources, fend off creatures, and construct tools.}
\label{fig:crafter-examples}
\end{figure}

\paragraph{Environment.} In the original Crafter environment\footnote{https://github.com/danijar/crafter}, agents are primarily tasked with exploration, gathering resources, and battling monsters. We have adapted this framework so that each episode now provides a free-form textual instruction, directing the agent to perform the number of specific tasks. In our modified version, each task given to the agent corresponds exactly to one achievement in the original game. Thus, within a single episode, following these instructions means completing individual achievements directly tied to the game's original goals. The examples of the agent in Crafter environment is presented in  Fig.~\ref{fig:crafter-examples}.

\paragraph{Dataset.} To create a dataset of textual tasks for the agent in the Crafter environment, we followed a specific pipeline: First, based on the known list of tasks and achievements in the crafter environment, we randomly generated a list of subtasks. Second, to obtain descriptive instructions for these tasks, we used Mistral-7B-Instruct\footnote{\href{https://huggingface.co/mistralai/Mistral-7B-Instruct-v0.1}{https://huggingface.co/mistralai/Mistral-7B-Instruct-v0.1}}. The example of such prompt and response is presented in Fig.\ref{fig:crafter-subtasks-to-text}. Third, to obtain more varied descriptions of instructions, using the same Mistral model to stylize the received instructions (for example of such reformulation see Fig.~\ref{fig:crafter-prompt-reformulation}).

\begin{figure}[ht]
    \centering
    \footnotesize
\begin{mdframed}[backgroundcolor=gray!10, roundcorner=10pt, linewidth=0pt, frametitle={\textbf{Example of prompt request:}}, frametitlebackgroundcolor=gray!20, nobreak=true]
    \footnotesize
    \noindent
    \texttt{Rewrite the list of subtasks into one instructive sentence for an agent in a virtual environment, replacing words from the list with synonyms. 
    \textit{Subtasks:} ["Defeat Zombie", "Collect Iron with count 1", "Make Iron Sword"]}
\end{mdframed}
\vspace*{-6px}

\begin{mdframed}[backgroundcolor=blue!10, roundcorner=10pt, linewidth=0pt, frametitle={\textbf{Example of response:}}, frametitlebackgroundcolor=blue!20, nobreak=true]
    \footnotesize
    \noindent
    \texttt{Vanquish the undead foe, gather a single unit of metallic mineral, and forge an iron weapon.}
\end{mdframed}
\vspace*{-5px}
    \caption{The example depict prompt and response interactions used to generate the dataset for the modified Crafter environment. }
    \label{fig:crafter-subtasks-to-text}
    \vspace*{-5px}
\end{figure}

\begin{figure}[ht]
\footnotesize
\begin{mdframed}[backgroundcolor=gray!10, roundcorner=10pt, linewidth=0pt, frametitle={\textbf{Example of prompt request:}}, frametitlebackgroundcolor=gray!20, nobreak=true]
    \footnotesize
    \noindent
    \texttt{Rewrite the instruction in the style of Tolkien with the same SHORT sentence size adding causality. 
    \textit{Instruction:} Vanquish the undead foe, gather a single unit of metallic mineral, and forge an iron weapon. }
\end{mdframed}
\vspace*{-6px}

\begin{mdframed}[backgroundcolor=blue!10, roundcorner=10pt, linewidth=0pt, frametitle={\textbf{Example of response:}}, frametitlebackgroundcolor=blue!20, nobreak=true]
    \noindent
    \texttt{Slay the wights, gather a nugget of mithril, and forge a blade of iron.}
\end{mdframed}
\vspace*{-5px}
\caption{This figure shows an examples prompt and response for the reformulation of an instruction generated by the Mistral-7B-Instruct.}
\label{fig:crafter-prompt-reformulation}

\end{figure}

\paragraph{Metrics.} To assess the quality of an agent's performance in an environment where it follows language instructions, we used a metric called "Success Rate." This metric measures whether all the subtasks required by the given instructions to the agent were completed. Accordingly, for one instruction, the metric can be $1$ if the agent completed all the tasks, or $0$ if any task was not completed. 

\paragraph{Baselines.} We compared our IGOR Agent with Dynalang~\cite{Dynalang}. To run Dynalang on a text-adapted Crafter, we employ the T5 model to encode the text, mirroring the approach taken by the paper's authors. In each episode, we sequentially transmit all instructions and images to Dynalang, one token at a time. This process allows us to evaluate Dynalang's ability to identify the necessary subtasks from the instructions, aligning with our methodology. We adopt the same reward system used in our experiments. 
 
\paragraph{Training details.} A \text{Language module} for the crafting environment was trained on the provided data without any modifications. Following the instructions within this environment, the module is tasked with determining which achievements the agent should collect in the virtual setting. Additionally, the Flan-T5-base model was utilized as a reference. 

For \text{Policy Module} we use self-supervised goal sampling to train the agent on these subtasks efficiently. In this approach, each training episode for the agent targets a distinct goal selected randomly, aligning with the goals possibly derived from the LLM. The reward function for the RL agent is designed to provide positive feedback for the successful completion of subtasks, motivating the agent to achieve its goals. This function uses both the original Crafter environment reward and an extra reward for the completion of a subtask by the agent. The original reward is scaled by a factor of $0.1$, and the extra reward for subtasks accomplishing is set to $1$. Through this process, the agent learns to solve individual subtasks, contributing to the overall action plan devised by the LLM.

\section{Experimental Results}

In this section, we compare our approach with other state-of-the-art methods. The section is organized as follows: First, we present the results in the modified Crafter environment. Second, we show the results for the IGLU environment, alongside ablation experiments for different data processing techniques.

\subsection{Crafter Environment}

\begin{figure}[htb!]
    \centering
    \includegraphics[width=1.0\linewidth]{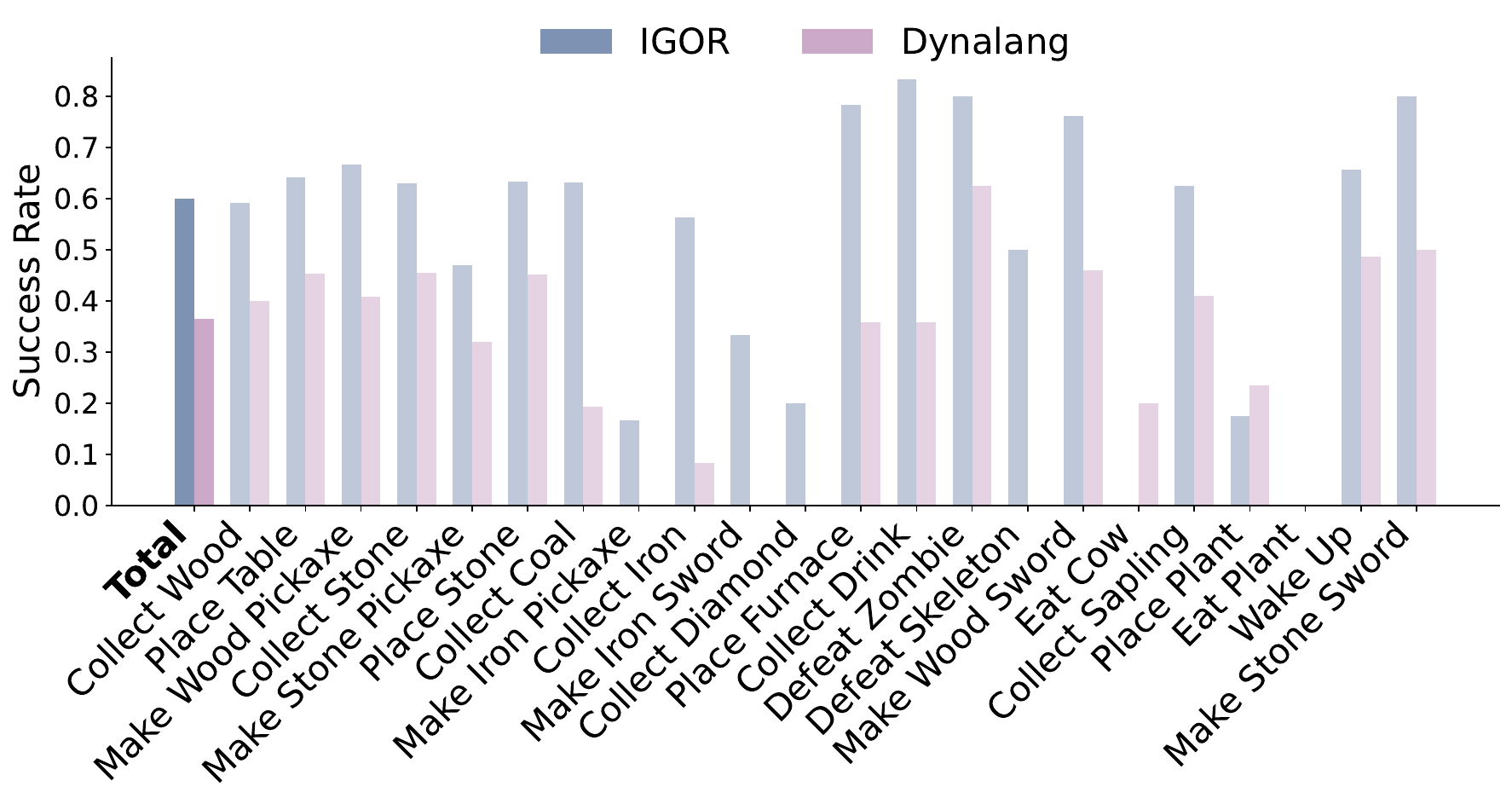}
    \vspace*{-15px}
    \caption{A comparison of the performance of IGOR and Dynalang approaches, on the Crafter environment with textual tasks. The success rate metric for each subtask is averaged across instructions requiring that specific subtask. \textit{Total} bar represents the overall success rate for all instructions of test dataset.}
    \label{fig:crafter_res}
\end{figure}

The results of IGOR and Dynalang in the Crafter environment are shown in the bar chart presented in Fig.~\ref{fig:crafter_res}. The x-axis displays various categories, where the \textbf{Total} category represents the overall success rate across all test instructions. The other categories illustrate the agents' performance on specific subtasks within the instructions, highlighting how effectively each agent solves individual subtasks when attempting to complete the entire instruction. For instance, a low value for a particular subtask on the plot indicates that the agent often fails to complete the instructions when that subtask is involved.

Overall, the total score of IGOR is $0.6$, indicating that the agent successfully completes all required subtasks in $60\%$ of cases, significantly outperforming Dynalang, which only achieves $36.4\%$. Our approach demonstrates superior performance in $19$ out of $22$ subtasks. Furthermore, it achieves results that are at least twice as good in $8$ out of $22$ types of instructions. Additionally, our approach, with the integration of curriculum learning, shows strong capabilities in managing tasks that are especially challenging, such as \texttt{Collect Diamond}, \texttt{Collect Iron}, \texttt{Make Iron Sword}, and \texttt{Make Iron Pickaxe}.

\subsection{IGLU Environemnt}

The comparative results in the IGLU environment are presented in Table~\ref{tab:iglu-results}. We ran the IGOR approach in both Flying and Walking modes. Additionally, we conducted experiments with different data variants for training the LLM agent, utilizing both primitives and coordinates as subtasks. Across all configurations, our method consistently outperforms the competitors, achieving significantly better results in all figure categories. Notably, IGOR’s lowest score is $0.45$, which surpasses the overall performance of the winning team, BrainAgent, which scored $0.36$. For simpler figures such as \texttt{Floor} using primitives, our agent scores $0.75$ compared to BrainAgent’s $0.53$. A clear advantage is also observed for more complex tasks such as \texttt{Air}, where IGOR scores $0.46$ versus $0.25$. Even when using coordinates as subtasks, our agent’s advantage, although reduced, persists across all figure types.

\begin{table}[ht!]

    \centering
    \fontsize{8}{9}\selectfont
    \caption{A comparative analysis of the F1 score (higher is better) for the IGOR approach versus the IGLU-2022 competition winners on the test dataset.}
    \label{tab:iglu-results}
    \begin{tabular}{lcc|cccc|c}
    \toprule
    Approach & Prim & Flying & Floor & Flat & Tall & Air & Total  \\
    \midrule
    IGOR & \checkmark & \checkmark & 0.72 & \textbf{0.51} & \textbf{0.46} & \textbf{0.40} & \textbf{0.52} \\
    IGOR & $\times$   & \checkmark & 0.68 & 0.44 & 0.36 & 0.34 & 0.45 \\

    IGOR & \checkmark & $\times$   & \textbf{0.75} & 0.46 & 0.32 &  0.33 & 0.46 \\
    IGOR & $\times$   & $\times$   & 0.68 & 0.42 & 0.29 & 0.31 & 0.45 \\
    
    \midrule 
    
    BrainAgent   & n\textbackslash a & $\times$ & 0.53 & 0.36 & 0.23 & 0.25 & 0.36 \\
    MHB          & n\textbackslash a & $\times$ & 0.08 & 0.05 & 0.03 & 0.04 & 0.05 \\
    Pegasus      & n\textbackslash a & $\times$ & 0.08 & 0.06 & 0.02 & 0.04 & 0.06 \\
    \bottomrule
    \end{tabular}
\end{table}

\paragraph{Comparing Data Processing Technics.} We evaluated various data processing strategies for training NLP models, as detailed in Table~\ref{llm_str_table}. These strategies included expanding the original dataset (strategy "Coords"), employing the ChatGPT augmentation technique outlined previously (strategy "ChatGPT"), and modifying the dataset's format (strategy "Prim"). We also enhanced the dataset with color augmentation by altering the color of the block and its corresponding descriptor in the instructions (strategy "Color").

\begin{table}[ht!]
\centering
\fontsize{8}{9}\selectfont
\caption{
    A comparative analysis of various data processing strategies for LLM training. The table displays the F1 scores, which measure discrepancy between LLM-predicted figures and target figures. The highest scores in each category are highlighted in bold.
    }
\label{llm_str_table}
\begin{tabular}
    {>{\centering\arraybackslash}p{0.5cm} >{\centering\arraybackslash}p{0.5cm} >{\centering\arraybackslash}p{1.0cm}|cccc|c}
    \toprule
    Prim & Color & ChatGPT & Floor & Flat & Tall & Air & Total \\
    
    \midrule
    \checkmark & \checkmark & \checkmark & \textbf{0.75} & 0.51 & \textbf{0.44} & \textbf{0.36} & \textbf{0.50} \\
    \checkmark & \checkmark &  $\times$          & 0.73 & \textbf{0.52} & 0.44 & 0.35 & \textbf{0.50} \\
    \checkmark & $\times$           &    $\times$        & 0.69 & 0.42 & 0.34 & 0.25 & 0.42 \\
    
    \midrule 
    $\times$ & \checkmark & \checkmark & 0.53 & 0.40 & 0.33 & 0.29 & 0.39 \\
    $\times$ & \checkmark &   $\times$         & 0.39 & 0.36 & 0.33 & 0.29 & 0.35 \\
    $\times$ &   $\times$         &    $\times$        & 0.45 & 0.35 & 0.30 & 0.28 & 0.34 \\
    \bottomrule
\end{tabular}
\end{table}

It can be observed that adaptation to primitives, even in the absence of augmentations, improves the F1 metric by $0.08$ (from $0.34$ to $0.42$). When combined with augmentations, the F1 score sees a $0.16$ boost ($0.50$) versus $0.01$ increase via augmentations alone on coordinates ($0.35$).

While the ChatGPT augmentations don't yield a significant increase on primitives, an analysis of the F1 scores across different shape categories reveals their effectiveness. On coordinates, the ChatGPT augmentation outperforms color augmentations, showing a $0.4$ relative improvement ($0.35$ vs $0.39$).

This suggests that the augmentation strategies have different impact depending on subtask format. Additionally, it is apparent that the primary challenge for the language model arises from instructions requiring the prediction of blocks suspended in the air—these tasks exhibit the highest error rates. This may be due to the limited number and specific nature of such instructions in the training dataset.

\section{Limitations}

Exploring previous studies in this area reveals that creating a general approach for multimodal environments presents a significant challenge~\cite{zhong2021silg}. While many approaches show potential, they often underperform when applied outside their original domains. Aiming for a universally applicable method is an admirable goal, yet it currently has limited practical use.

Our work introduces a new framework that separates the training of RL and LLM components. The proposed decomposition aims to achieve architectural flexibility, facilitating the easy integration of various techniques such as curriculum learning and reward shaping for RL, as well as data augmentation and human feedback for LLM training. However, our approach requires the identification of known subtasks, which may not be present in some domains, thus necessitating additional data collection to identify them.

\section{Conclusion}

In this paper, we introduce the Instruction Following with Goal-Conditioned Reinforcement Learning in Virtual Environments (IGOR) method, a novel approach that translates natural language instructions into a sequence of executable subtasks. The IGOR method integrates two independently trainable modules along with an intermediary Task Manager module. The Language Module is responsible for converting textual instructions into a defined sequence of subtasks. Meanwhile, the RL agent is structured in a goal-conditioned manner, making it adept at handling a wide array of tasks across various environmental contexts.

The modular decomposition of the IGOR approach allows for the incorporation of additional training techniques such as data augmentation, goal-based strategies, and curriculum learning in RL. A detailed analysis of the results shows that this approach not only allows for more flexible training customization but also yields a significant improvement in performance.

We demonstrate these advantages through experiments in two environments. In the IGLU environment, where the agent is required to construct a figure in a virtual setting based on dialogue with a human, our method surpasses the winners of the IGLU 2022 competition. Additionally, we have adapted the Crafter environment to require the agent to achieve specific achievements based on instructions, showing that our approach outperforms the Dynalang method based on Dreamer V3.
 \vspace*{-5px}

\bibliography{bib}

\newpage

\appendix


\section{Curriculum Learning}


We use curriculum learning technique, inspired by principles outlined in \cite{matiisen2019teacher, nesterova2022reinforcement}. The core objective of this method is to dynamically adjust the selection probabilities of training tasks to optimize learning progression based on empirical performance measures. The pseudocode is outlined in Algorithm~\ref{algo:curriculum}.

Our algorithm commences with an equal likelihood of selecting any given task from a list of tasks $\mathcal{T}$, each characterized by two parameters: an average reward $r_i$, computed using an exponential moving average, and a reward variability measure $\delta_i = |r_i^t - r_i^{t-1}|$, which captures the absolute change in rewards between consecutive trials. This dynamic adjustment is governed by two key parameters: a scaling coefficient $d$ and a success rate threshold $\tau$.

The task selection probability is modified according to the task's performance relative to $\tau$. For tasks where the reward $r_i$ meets or exceeds $\tau$, the algorithm assigns a lower selection probability of $q_i = \frac{1}{d}$, effectively deprioritizing tasks that have already surpassed a certain success benchmark. Conversely, tasks that fall below this threshold are given a higher probability of selection, formulated as $q_i = 1 + (\delta_i \cdot d)$. This scheme favors tasks with greater variability in performance, hypothesizing that such tasks may yield more informative learning experiences due to their higher potential for improvement.

To ensure a probabilistic selection, the probabilities are normalized using a softmax function, yielding a distribution $\mathbf{p}$. A task is then selected randomly according to this distribution, with the likelihood of selecting task $i$ given by $p_i$ from $\mathbf{p}$, effectively integrating an element of stochasticity into task selection which is critical for exploring a variety of learning experiences. This methodology allows for a balanced exploration-exploitation trade-off, adapting task difficulty based on learner performance to potentially accelerate skill acquisition in complex learning environments.

This adaptive approach to task selection represents a significant enhancement to the efficiency of curriculum learning strategies, ensuring that learners are consistently challenged with tasks appropriate to their evolving capabilities. This strategy not only maintains engagement but also maximizes learning outcomes by strategically modulating the difficulty of tasks presented over the course of training.

\begin{algorithm}[ht!]
\caption{Task Selection Based on Success Rate}
\label{algo:curriculum}

\KwIn{List of tasks $\mathcal{T}$, each task $i$ characterized by parameters $(r_i, \delta_i)$ where $r_i$ is the average reward for task $i$ using exponential moving average, and $\delta_i = |r_i^t - r_i^{t-1}|$ is the average difference between subsequent rewards for task $i$, also using exponential moving average. Constants $d$ (a coefficient that influences sampling probability) and $\tau$ (a success rate threshold).}
\KwOut{Selected task}

\SetKwInput{KwInput}{Input}
\SetKwInput{KwOutput}{Output}

\SetAlgoLined
\SetKwProg{Fn}{Function}{:}{end}
\SetKwFunction{FAdaptiveSelect}{AdaptiveSelect}

\Fn{\FAdaptiveSelect{$\mathcal{T}, d, \tau$}}{
    Initialize a vector $\mathbf{q} \leftarrow []$\;
    \For{each task $i \in \mathcal{T}$}{
        \eIf{$r_i \geq \tau$}{
            \tcp{Assign a lower probability for tasks that exceed the success threshold $\tau$}
            $q_i \leftarrow \frac{1}{d}$\;
        }{
        \tcp{Assign higher probability to tasks with more variability and below the threshold}
            $q_i \leftarrow 1 + (\delta_i \cdot d)$\;
            
        }
        Append $q_i$ to $\mathbf{q}$\;
    }
    \tcp{Normalize probabilities}
    $\mathbf{p} \leftarrow \text{Softmax}(\mathbf{q})$\;

    \tcp{Sample task index $j$ according to the probability distribution $\mathbf{p}$}
    $j \sim \text{Categorical}(\mathbf{p})$\;

    \KwRet $\mathcal{T}[j]$\;
}
\end{algorithm}

\begin{figure*}
    \centering
    \includegraphics[width=0.85\linewidth]{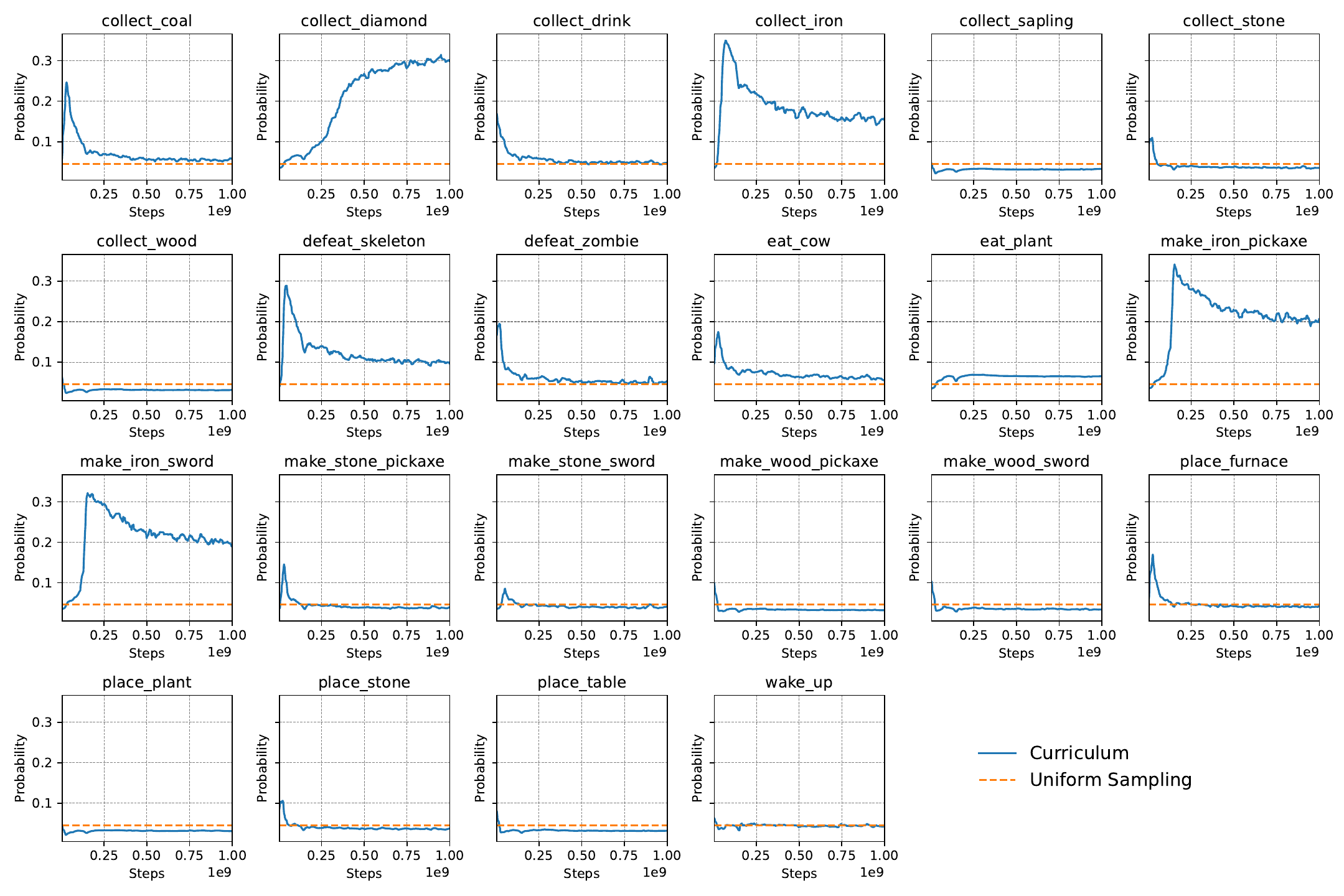}
    \caption{Comparison of probabilities of sampling a subtask during training between a uniform sampling policy and a curriculum learning-based policy.}
    \label{fig:plot:probs}
\end{figure*}

\begin{figure*}
    \centering
    \includegraphics[width=0.85\linewidth]{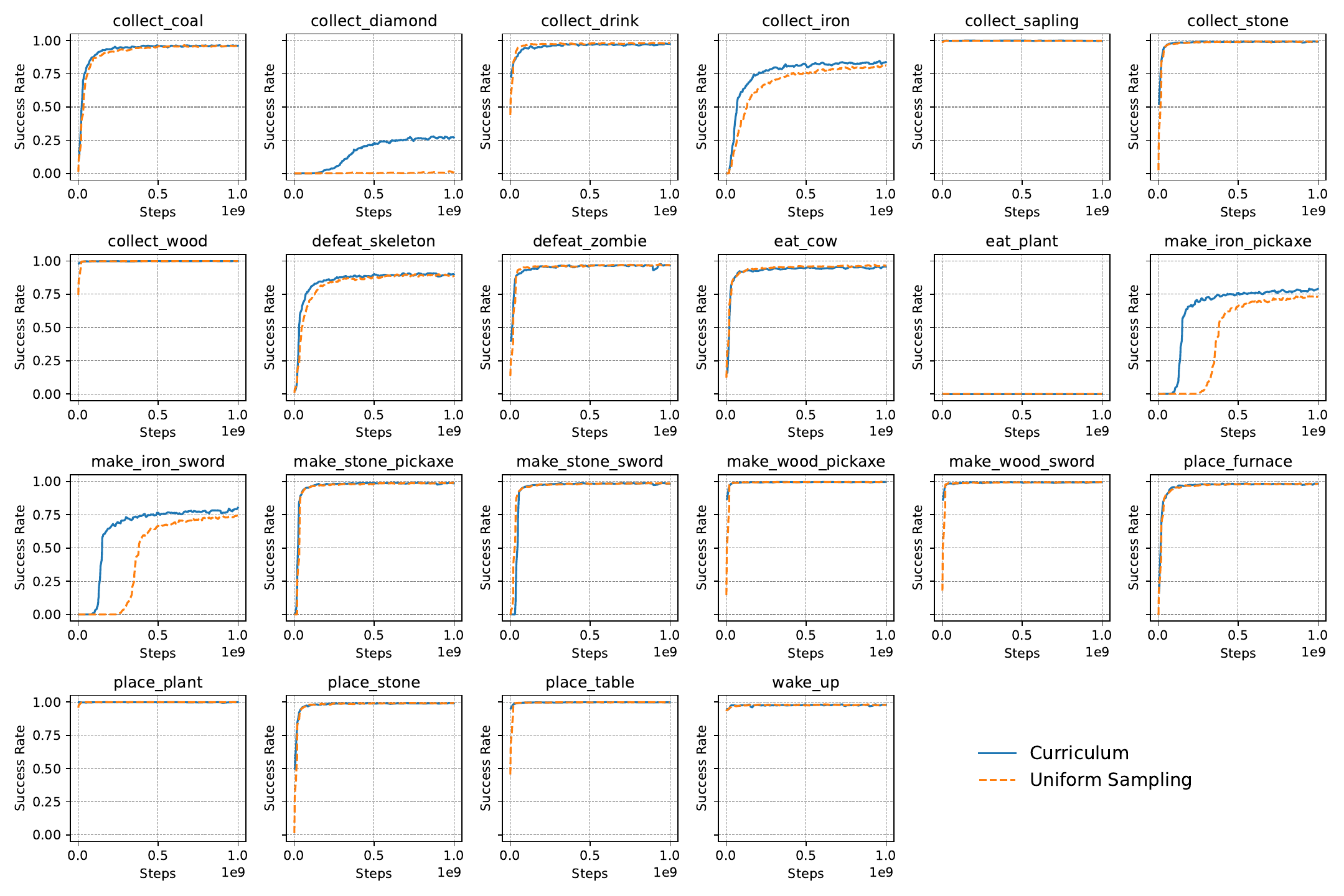}
    \caption{Success rates for each subtask during training, comparing a uniform sampling policy to a curriculum learning-based policy.}
    \label{fig:plot:sr}
\end{figure*}

To highlight the importance of curriculum learning techniques, we conducted an additional experiment in the Crafter environment. Here, we compare the learning processes of a PPO agent with and without curriculum learning. In Figure~\ref{fig:plot:probs}, we show the probability of sampling each subtask during training. For the uniform policy, this is represented by a line parallel to the x-axis. In Figure~\ref{fig:plot:sr}, we provide the success rates for both approaches. It is clear that the version with curriculum learning shows better results for tasks such as \textit{collect\_iron}, \textit{collect\_diamond}, \textit{make\_iron\_pickaxe}, and \textit{make\_iron\_sword}. This improvement can be attributed to the dynamics shown in the first plot. For example, for the \textit{collect\_iron} task, as the agent begins to master this problem, the sampling probability increases. Once performance plateaus, the probability decreases. This pattern also applies to other tasks; if the agent has already mastered a subtask, e.g., \textit{place\_table} or \textit{make\_wood\_pickaxe}, the probability of sampling these subtasks decreases below that of uniform sampling, thereby freeing up resources for learning other subtasks.

\section{Coordinates Generation Prompt }\label{ap-gen-prompt}

 First, we include a description of the IGLU environment in the prompt. This way, we provide information on how the coordinates change in relation to cardinal directions (north, west, sky). We also provide information about certain patterns present in the dataset (rows, squares, columns, towers) and what they signify in the dataset.

 \textit{Environment have size (11 11 9). Coordinates are (x y z). x coordinate increases towards south. y coordinate increases towards east. z coordinate increases towards the sky. So the highest south-west coordinate would be ( 11 11 9). And the lowest nort-west coordinate is (0 0 0). Middle of enviroment/grid is (5 5 0). If I will start from the middle of enviroment/grid, but on the floor, face north and put one block, it coords will be (4 50). I can place blocks in this environment. For task: "Facing north, build three stacks of two orange blocks, then destroy the bottom orange block on all three stacks"; I will get a cords for built figure: ((2 4 0), (2 5 0), (2 6 0)). If I want a column/tower of 6 yellow blocks in the middle of the grid, the coords will be (5 5 0), (5 5 1), (5 5 2), (5 5 3), ( 5 5 4), (5 5 5). If I want a row with 3 blocks towards south: (5 5 0), (5 6 0), (5 7 0). What coords will be for tower of 6 yellow blocks in the middle of the grid? As aswer, send me only list of coords.}

Next, we ask to add the color of the shape to the coordinate description, which should be chosen for a specific block.

\textit{As aswer, send me only list of coords. Add color digit to tuples array with the next color mapping: BLUE - 1 GREEN - 2 RED - 3 ORANGE - 4 PURPLE - 5 YELLOW - 6 and use 0 - for blocks needs to be removed}

Now, in the chatbot's single-session mode, we sequentially present the instructions from the task without making any alterations to them.

\section{Instruction Augmentation Prompt }\label{ap-aug-prompt}

 In this prompt, we also start by providing information about the environment like in \ref{ap-gen-prompt} and the dialogues. Then, we ask to rewrite the dialogue as if the shapes were required to be built rotated by 180 degrees.

\textit{I have a dialog with instructions describing how to build a 3D figure on a plane. The instructions have information on where to build the shapes and how to build them. I will consistently give you dialogues, and you have to rewrite the instructions so that they describe the construction of the same figure, only the location is rotated 180 degrees clockwise. 
Size of eviroment is (9, 11, 11)
Remember: column and tower are vertical structures, line and row are horizontal.
Do not change the instructions associated with the modification of vertical structures (keywords top and bottom).
The output is only the final dialogue (be sure to save the ++ signs, I will then use them to separate the dialogue into instructions). }

\section{IGLU Structure Classes}\label{iglu_figs}

\begin{figure}[htb!]
    \centering
    \includegraphics[width=0.87\linewidth]{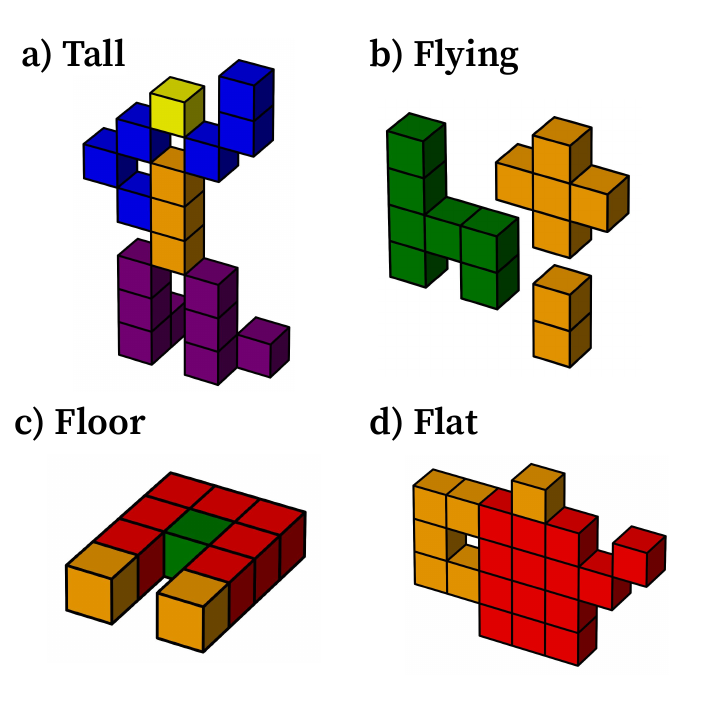}
    \vspace*{-8px}
    \caption{Example of IGLU figures classes.}
    \label{fig:appendix:iglu-structrues}
\end{figure}

Examples of structure classes in the IGLU environment are presented in Fig.~\ref{fig:appendix:iglu-structrues}. Each structure represents a typical example used in the experimental results section. Please note that the same figure can be attributed to several classes; for example, structure (a) is both \texttt{tall} and \texttt{flying} simultaneously. The structure on the \texttt{floor} (c) is also a flat structure, etc.

The complexity of each class is highlighted by the experimental results. \texttt{tall} and \texttt{flying} structures are more difficult than \texttt{floor} and \texttt{flat} structures. \texttt{floor} is the easiest one, since it doesn't require the ability to place or remove proper supporting blocks, which are much more challenging tasks.

\section{IGLU Reward Function}

In the IGLU environment, the reward function used changes throughout the training curriculum to balance guided exploration and precise task execution. Initially, the primary reward is based on the distance from the modification to the goal, calculated as $\frac{+1}{dist + 1}$. As training progresses, this reward structure transitions to a more rigorous system where the agent receives $+1$ for correct modifications (indicating successful subtask completion) and a constant penalty of $-0.5$ for incorrect modifications, regardless of the distance. Additionally, in the early stages of the curriculum, there is an auxiliary reward for reducing the distance between the agent and the goal, provided only when moving positively towards the goal. This proximity reward diminishes as training advances, encouraging the agent to autonomously navigate and execute tasks by the end of the curriculum.

\section{Hyperparameters and Hardware Resources}\label{ap-params}

Table~\ref{table:parameters} lists the hyperparameters used for training the RL module of the IGOR approach. Table~\ref{table:flant5_parameters} provides the hyperparameters for training the Flan T5 language model, specifying only those parameters that were adjusted from the default settings in the HuggingFace.

\begin{table}[htb!]
    \centering
    \caption{The parameters utilized for training the PPO actor-critic model.
    }
    \small
    \label{table:parameters}
    \begin{tabular}{ll}

        \toprule
        Hyperparameter            & Value             \\
        \midrule
        Adam learning rate        & $0.0001$        \\
        $\gamma$  (discount factor) & $0.99$        \\
        Rollout                   & $32$               \\
        Clip ratio                & $0.1$        \\
        Batch size                & $1024$            \\
        Optimization epochs       & $1$               \\
        Entropy coefficient       & $0.003$        \\
        Value loss coefficient    & $0.5$             \\
        GAE$_\lambda$             & $0.95$            \\
        \midrule
        ResNet residual blocks    & 3                 \\
        ResNet number of filters  & $64$              \\
        LSTM hidden size          & $512$             \\
        Activation function       & ReLU              \\
        Network Initialization    & orthogonal        \\
        Rollout workers           & $16$               \\
        Environments per worker   & $32$               \\
        \bottomrule
    \end{tabular}
    \end{table}

\begin{table}[htb!]
    \centering
    \caption{Parameters used to finetune Flan-T5, which have been changed from the default parameters in the Transformers library.
    }
    \small
    \label{table:flant5_parameters}
    \begin{tabular}{ll}

        \toprule
        Hyperparameter            & Value             \\
        \midrule
        learning rate        & $0.0001$        \\
        batch size       & $16$        \\
        gradient accumulation steps      & $16$        \\
        \bottomrule
    \end{tabular}
    \end{table}

Table~\ref{tab:training_times_devices} shows the training time and hardware resources. IGOR uses 139 GPU hours in Crafter and 32 in IGLU, while Dynalang requires 166 GPU hours in Crafter. IGOR efficiently combines RL and LLM modules on various devices.

\begin{table}[ht!]
\centering
\caption{Training time and hardware resources for different modules}
\label{tab:training_times_devices}
\begin{tabular}{lrc}
\toprule
Training Module & GPU hours & Device \\ 
\midrule
Dynalang (crafter) & 166 & TITAN RTX \\ 
IGOR (RL crafter) & 136 & TITAN RTX \\ 
IGOR (LLM crafter) & 3 & NVIDIA A100 \\
\midrule
IGOR (RL iglu)  & 20  & TITAN RTX \\ 
IGOR (LLM iglu) & 12  & NVIDIA A100  \\ 
\bottomrule
\end{tabular}

\end{table}

\end{document}